\documentclass[letterpaper]{article} 
\usepackage{aaai2026}  
\usepackage{times}  
\usepackage{helvet}  
\usepackage{courier}  
\usepackage[hyphens]{url}  
\usepackage{graphicx} 
\usepackage{natbib}  
\usepackage{caption} 
\usepackage{amsmath} 
\usepackage{amsmath, amssymb, amsfonts}
\usepackage{xcolor}
\usepackage{amstext}
\usepackage{multirow, booktabs} 
\usepackage{makecell}
\usepackage{amsthm,mathtools}
\usepackage{thmtools}
\usepackage{subfigure}

\usepackage{algorithm}
\usepackage{algorithmic}

\definecolor{mblue}{HTML}{86addc}     
\definecolor{mpurple}{HTML}{9375d8}  
\usepackage{newfloat}
\usepackage{listings}
\DeclareCaptionStyle{ruled}{labelfont=normalfont,labelsep=colon,strut=off} 
\floatstyle{ruled}
\newfloat{listing}{tb}{lst}{}
\floatname{listing}{Listing}
\title{Memory-Based Advantage Shaping for LLM-Guided Reinforcement Learning}

\author{
  Narjes Nourzad\textsuperscript{\rm 1}\thanks{This work was conducted during an internship at Carnegie Mellon University.},
  Carlee Joe-Wong\textsuperscript{\rm 2}
}

\affiliations{
  \textsuperscript{\rm 1}University of Southern California\\
  \textsuperscript{\rm 2}Carnegie Mellon University\\
  nourzad@usc.edu, cjoewong@andrew.cmu.edu
}

\usepackage{bibentry}

\begin{document}

\maketitle

\begin{abstract}

In environments with sparse or delayed rewards, reinforcement learning (RL) incurs high sample complexity due to the large number of interactions needed for learning.
This limitation has motivated the use of large language models (LLMs) for subgoal discovery and trajectory guidance.
While LLMs can support exploration, frequent reliance on LLM calls raises concerns about scalability and reliability.
We address these challenges by constructing a \textit{memory graph} that encodes subgoals and trajectories from both LLM guidance and the agent’s own successful rollouts. 
From this graph, we derive a \textit{utility function} that evaluates how closely the agent’s trajectories align with prior successful strategies. 
This utility \textit{shapes the advantage function},  providing the critic with additional guidance without altering the reward.
Our method relies primarily on offline input and only occasional online queries, avoiding dependence on continuous LLM supervision. Preliminary experiments in benchmark environments show improved sample efficiency and faster early learning compared to baseline RL methods, with final returns comparable to methods that require frequent LLM interaction.

\end{abstract}


\section{Introduction}

Reinforcement learning (RL) has achieved state-of-the-art results across diverse domains~\cite{nourzadactor, liu2024integrating}, yet its limitations become evident in settings that more closely resemble real-world conditions. 
Sparse or delayed rewards, partial observability, and long decision horizons often result in poor sample efficiency and unstable training dynamics.
Large language models (LLMs) can help address these issues by providing high-level priors, suggesting subgoals, and structuring exploration~\citep{schoepp2025evolving}. 
At the same time, LLMs introduce new challenges. 
Their outputs may be unreliable, queries can be expensive, and continuous supervision risks diluting the role of environment-driven feedback.
Thus, it is challenging to integrate LLM guidance in a way that enhances RL without creating excessive dependence. The goal is to ensure external signals complement rather than distort optimization.

In this work, we introduce an approach that incorporates external guidance into RL models through \textit{advantage shaping}. 
Central to the method is a \textit{memory graph} co-constructed from the agent’s rollouts together with offline priors and infrequent online outputs from an LLM. 
The graph provides a compact representation of subgoals and task-relevant knowledge that evolves during training, reducing dependence on real-time LLM access.
A utility signal derived from the memory graph is used to softly shape advantage estimates, guiding policy updates and improving early exploration. 
The influence of this shaping term becomes asymptotically negligible once the agent’s policy surpasses the usefulness of LLM-derived guidance, ensuring that the convergence guarantees of PPO (proximal policy optimization)~\citep{schulman2017ppo} remain intact.

 \section{Methodology}
 \begin{figure}
     \centering
     \includegraphics[width=0.6\linewidth]{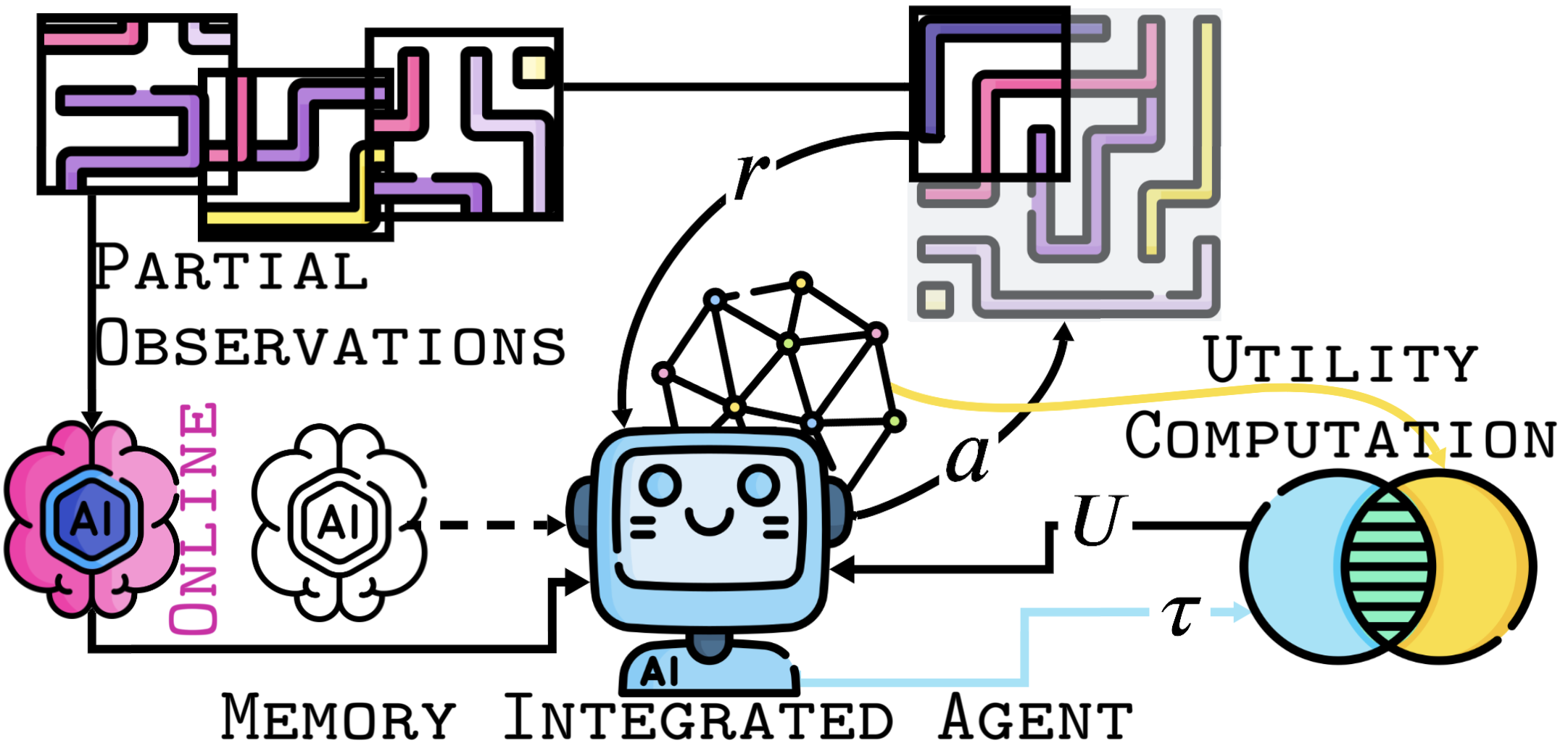}
     \caption{Overview of the proposed method.}
     \label{fig:main}
 \end{figure}
 Our method integrates external guidance through a memory graph, initialized with offline LLM priors (e.g., partial trajectory and subgoal decompositions) and augmented with agent rollouts and occasional online queries. Online queries are triggered adaptively when the agent fails to extract useful guidance from memory for several consecutive episodes. The LLM is constrained to the same partial observability as the agent and returns short plans that are either added to the graph or used to bias action preferences through soft logit injection. Formally, the memory graph is given by
$$\mathcal{G} = 
\Bigl\{ \big((o, a)_{\tau_j}, \zeta_j, \hat r_j\big)\Bigr\}_{j=1}^N 
\;\cup\; \bigl\{ \kappa_\ell \bigr\}_{\ell=1}^L 
\;\cup\; \{ \textsl{g}_\triangleright \}. $$
Each trajectory node $j$ consists of a partial observation $o_{\tau_j}$, action $a_{\tau_j}$, goal term $\zeta_j \in \{\textsl{g}_j,\, \kappa_{\ell}^{\textsl{g}_j}\}$ indicating a final goal $(\textsl{g}_j)$ or an abstract subgoal $(\kappa_{\ell}^{\textsl{g}_j})$, and estimated reward $\hat r_j$ for the corresponding action sequence. The second set of nodes $\{\kappa_\ell\}_{\ell=1}^L$ represents subgoals inferred from environment descriptions, while the final term $\{\textsl{g}_\triangleright\}$ denotes the agent’s target goal(s).  These nodes are connected through goal–subgoal relationships provided directly by the LLM.
The graph evolves during training, with new nodes added when novel behaviors are observed and rarely accessed nodes pruned as obsolete to stay keep the structure compact while ensuring consistent guidance with little LLM reliance.


For each state-action pair $(o_t, a_t)$, utility is computed by evaluating the agent’s observed behavior against the memory graph. It leverages the same rollouts used for advantage estimation under the current policy. 
Each pair in the trajectory $\tau = {(o_t, a_t)}_{t=1}^T$ is compared to a stored trajectory node ($m$) from the memory graph,  aligned with the environment layout in the current rollout. Formally, 
$U_t \doteq  \hat{r}_{m} \cdot \rho(\textsl{g}_\triangleright, \zeta_{m}) \cdot \mathcal{s}\big((o_t,a_t), (o_{t’},a_{t’})_{\tau_m}\big).$
The similarity term $\mathcal{s}(\cdot,\cdot)$ measures action agreement and how closely the agent’s trajectory follows the path implied by the stored segment (e.g., position or directional overlap). 
The factor $\rho(\cdot,\cdot)$ serves to downweight matches that are behaviorally similar but pursue different subgoals, computed as the Jaccard similarity between the goals in the rollout and those encoded in memory.  
This ensures that utility reflects both behavioral similarity and semantic alignment with successful prior strategies. 
Overview of the  method in Fig.~\ref{fig:main}.

\begin{algorithm}
\caption{Shaped PPO actor (\textcolor{mpurple}{changes in purple})}
\label{alg:ppo_ours}
\begin{algorithmic}
\STATE Collect  $\mathcal{D} = \{(s_t, a_t, r_t)\}$ using $\pi_{\theta}$
\STATE Compute ${A}_t$ and \textcolor{mpurple}{$U_t$} from rollouts \label{line:adv_shaping}
\STATE ${\textcolor{mpurple}{\tilde{A}_t}}= {A}_t + \textcolor{mpurple}{\xi_t U_t}$
\FOR{$epoch = 1$ to $K$, minibatch $\mathcal{B} \subset \mathcal{D}_k$}
        \STATE $r_t (\theta) = {\pi_\theta(a_t|s_t)}/{\pi_{\theta_k}(a_t|s_t)}$
        \STATE \textcolor{mpurple}{$\mathcal L^{\text{shaped}}(\pi_\theta)$} $= \mathbb{E}\left[\min(r_t,1\!\pm\!\varepsilon_k){\textcolor{mpurple}{\tilde{A}_t}} \right]$
        \STATE $\theta \leftarrow \theta + \alpha_\theta \nabla_\theta$ \textcolor{mpurple}{$\mathcal L^{\text{shaped}}(\pi_\theta)$}
\ENDFOR
\end{algorithmic}
\end{algorithm}

We incorporate memory-derived utility into the policy update by augmenting the standard advantage term. The advantage function quantifies how favorable an action is relative to the average action at state, reinforcing those with higher-than-expected returns and suppressing those that fall short. However, during early training the critic is poorly calibrated due to limited exploration, often producing near-uniform or noisy value estimates~\cite{henderson2018deep}. As a result, the computed advantages provide weak learning signals, even when the behavior is directed toward the task, leading to inefficient or unstable updates. This issue is more pronounced in sparse-reward or delayed-feedback settings, where $A_t$ tends to be close to zero for most initial training timesteps. To address this, we define a \textit{shaped advantage} 
$\tilde{A}_t = A_t + \xi_t U_t \quad \text{where } 0 < \xi_t \leq 1 .$
When critic feedback is weak, $U_t$ supplies additional direction aligned with task objectives, compensating for flat or noisy gradients and accelerating learning. As training progresses and $A_t$ becomes more reliable, the contribution of $U_t$ naturally becomes negligible. Because the utility enters additively and does not modify the reward signal, the optimization dynamics of PPO are preserved, along with its convergence guarantees. More broadly, this mechanism is compatible with policy gradient methods relying on advantage estimation, providing a general way to incorporate structured guidance into RL.

\begin{figure}[t]
    \centering
    \includegraphics[width=.8\linewidth]{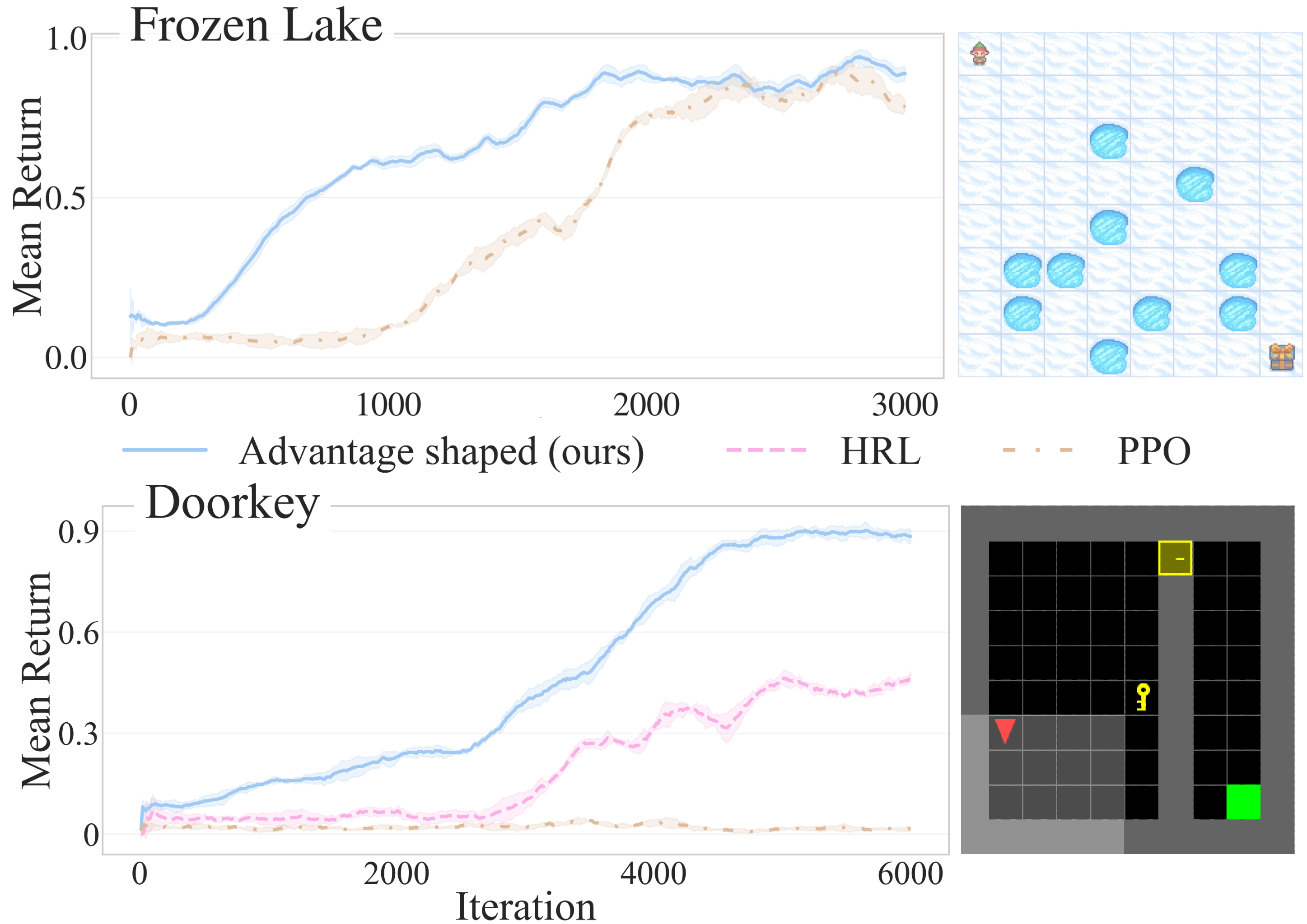}
    \caption{Mean return on \textsc{FrozenLake-8x8} (top) and \textsc{Doorkey} (bottom). Our method has the highest return.}
    \label{fig:main_shaped}
\end{figure}


\section{Experimental Results}

\subsubsection{Experimental Setup}
We evaluate our method in two environments: \textsc{FrozenLake} (Gymnasium) and \textsc{Doorkey} (MiniGrid suit), selected to probe early-stage exploration effects and long-horizon, partially observable planning, respectively. 
We compare against three baselines: PPO, trained tabula rasa from rewards, Hierarchical RL (HRL)~\citep{bhambri2024extracting}, which uses LLM-derived option policies for temporal abstraction, and  LLM4Teach~\cite{zhou2023large}, a state-of-the-art method where a pre-trained LLM distills guidance into a student RL agent.

Figure~\ref{fig:main_shaped} reports mean return across environments. 
In \textsc{FrozenLake}, shaped advantages achieve faster convergence, but baselines eventually reach the same asymptotic return with sufficient training.
In \textsc{DoorKey}, however, the shaped agent sustains a clear advantage, achieving both higher sample efficiency and stronger final performance. While HRL improves over PPO,  it trails our method by nearly a factor of two.
Table~\ref{tab:doorkey} reports test-time evaluation on unseen seeds for \textsc{DoorKey}. 
Here, our method matches the state-of-the-art LLM4Teach in both mean return and success rate, with no statistically significant difference at the $95\%$ confidence level. 
Unlike teacher-based approaches, however, these gains are achieved without continuous LLM supervision, relying only on limited offline and occasional online queries.
\begin{table}[h]
\centering
\begin{tabular}{lcc}
\toprule
\textbf{Method} & Mean Return & Success Rate \\
\midrule
LLM4teach   & $0.912 \pm 0.075$ & $0.970 \pm 0.004$ \\
\textbf{Our method}        & $0.898 \pm 0.093$ & $0.953 \pm 0.043$ \\
\bottomrule
\end{tabular}
\caption{Performance comparison on unseen seeds.}
\label{tab:doorkey}
\end{table}
 These results indicate that memory-derived utility provides meaningful guidance in sparse and partially observable environments while minimizing the reliance on expensive LLM queries.

{
\bibliography{aaai2026}}

\end{document}